\documentclass[10pt,twocolumn,letterpaper]{article}

\usepackage{cvpr}              % To produce the CAMERA-READY version

\usepackage[accsupp]{axessibility}
\usepackage{graphicx}
\usepackage{amsmath}
\usepackage{amssymb}
\usepackage{booktabs}
\usepackage[linesnumbered,ruled]{algorithm2e}
\usepackage{xcolor}
\usepackage{amsmath}
\usepackage{subcaption}
\usepackage{multirow}
\usepackage{pifont}% http://ctan.org/pkg/pifont
\usepackage{enumitem}

\usepackage[pagebackref,breaklinks,colorlinks]{hyperref}

\usepackage[capitalize]{cleveref}
\crefname{section}{Sec.}{Secs.}
\Crefname{section}{Section}{Sections}
\Crefname{table}{Table}{Tables}
\crefname{table}{Tab.}{Tabs.}

\usepackage{xcolor}
\usepackage{circledsteps}
\newcommand\myCircled[2][]{\ifmmode
\Circled[fill color=black,inner color=white,#1]{\mathsf{#2}}
\else
\Circled[fill color=black,inner color=white,#1]{\sffamily#2}
\fi
}

\def\cvprPaperID{7043} % *** Enter the CVPR Paper ID here
\def\confName{CVPR}
\def\confYear{2023}

\usepackage{xr}
\makeatletter

\newcommand*{\addFileDependency}[1]{% argument=file name and extension
\typeout{(#1)}% latexmk will find this if $recorder=0
\@addtofilelist{#1}
\IfFileExists{#1}{}{\typeout{No file #1.}}
}\makeatother

\newcommand*{\myexternaldocument}[1]{%
\externaldocument{#1}%
\addFileDependency{#1.tex}%
\addFileDependency{#1.aux}%
}
\myexternaldocument{Supplementary} % not working somehow

\begin{document}

\title{PaCa-ViT: Learning Patch-to-Cluster Attention in Vision Transformers}

\author{Ryan Grainger$^1$, Thomas Paniagua$^1$, Xi Song$^2$, Naresh Cuntoor$^3$, Mun Wai Lee$^3$ and Tianfu Wu$^1$\thanks{T. Wu is the corresponding author.}\\
$^1$Department of ECE, NC State, \quad 
$^2$An Independent Researcher,  \quad  
$^3$BlueHalo \\
% {\tt\small \{rpgraing, tapaniag, twu19\}@ncsu.edu, xsong.lhi@gmail.com,} \\{\tt\small \{Naresh.Cuntoor,MunWai.Lee\}@bluehalo.com}
% \and
}
\maketitle

%%%%%%%%% ABSTRACT
\begin{abstract}
Vision Transformers (ViTs) are built on the assumption of treating image patches as ``visual tokens" and learn patch-to-patch attention. The patch embedding based tokenizer has a semantic gap with respect to its counterpart, the textual tokenizer. The patch-to-patch attention suffers from the quadratic complexity issue, and also makes it non-trivial to explain learned ViTs. To address these issues in ViT, this paper proposes to learn Patch-to-Cluster attention (PaCa) in ViT. Queries in our PaCa-ViT starts with patches, while keys and values are directly based on clustering (with a predefined small number of clusters). The clusters are learned end-to-end, leading to better tokenizers and inducing joint clustering-for-attention and attention-for-clustering for better and interpretable models. The quadratic complexity is relaxed to linear complexity. The proposed PaCa module is used in designing efficient and interpretable ViT backbones and semantic segmentation head networks. In experiments, the proposed methods are tested on ImageNet-1k image classification, MS-COCO object detection and instance segmentation and MIT-ADE20k semantic segmentation. Compared with the prior art, it obtains better performance in all the three benchmarks than the SWin~\cite{liu2021swin} and the PVTs~\cite{wang2021pyramid,wang2021pvtv2} by significant margins in ImageNet-1k and MIT-ADE20k. It is also significantly more efficient than PVT models in MS-COCO and MIT-ADE20k due to the linear complexity. The learned clusters are semantically meaningful. 
Code and model checkpoints are available at \url{https://github.com/iVMCL/PaCaViT}.
\end{abstract}

% figure
\setlength{\abovecaptionskip}{0pt}
\setlength{\belowcaptionskip}{-6pt}
% equation
\setlength{\belowdisplayskip}{1pt} \setlength{\belowdisplayshortskip}{1pt}
\setlength{\abovedisplayskip}{1pt} \setlength{\abovedisplayshortskip}{1pt} 

\section{Introduction}\label{sec:intro}
\vspace{-2mm}
\begin{figure}[t]
  \centering  
  \includegraphics[width=0.48\textwidth]{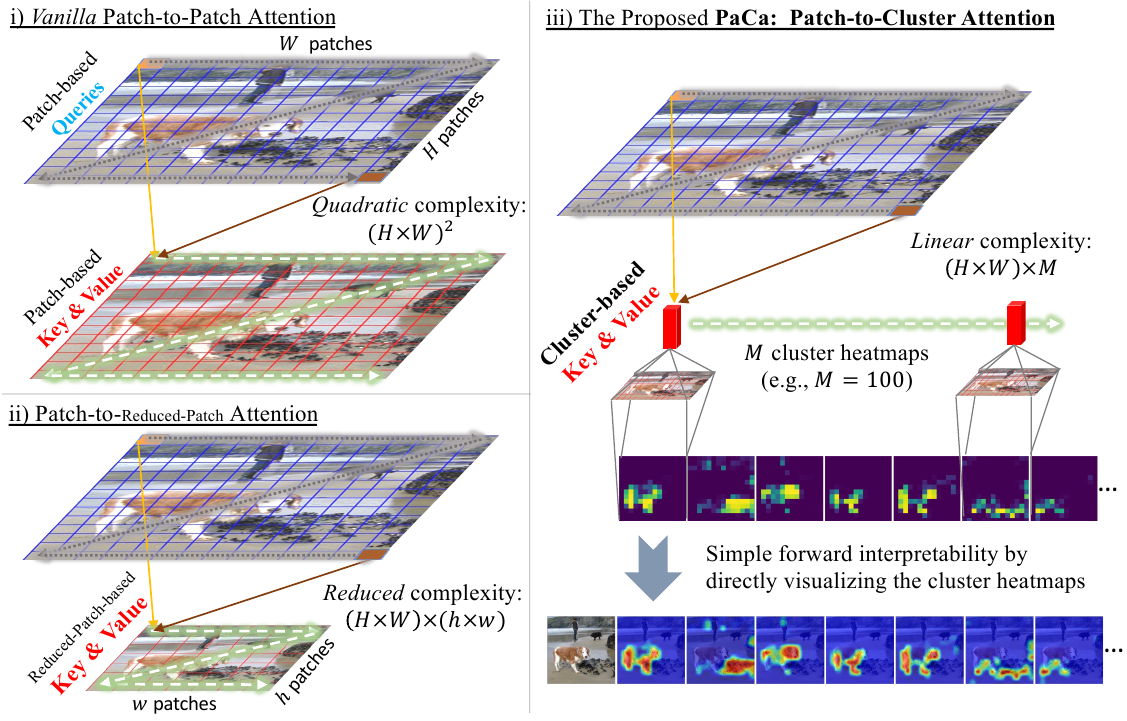}
  \vspace{-4mm}
  \caption{\small i) The vanilla patch-to-patch self-attention~\cite{vaswani2017attention,dosovitskiy2020image} directly leverages image patch embeddings as visual tokens and suffers from its quadratic complexity. Every Query (e.g., the patches in the blue grid) needs to interact with every Key. ii) To address the quadratic complexity, one popular method is to leverage  spatial reduction (e.g., implemented via a convolution with a stride $r>1$) in computing the Key and the Value~\cite{wang2021pyramid,wang2021pvtv2}. It still performs patch-to-patch attention, but enjoys a reduced complexity. iii) We propose \textbf{Patch-to-Cluster attention (PaCa)} in this paper. A predefined number of $M$ cluster assignments is first learned and then used in computing the Key and Value, resulting in not only linear complexity, but also more meaningful visual tokens. %The $M$ cluster heatmaps can either be computed from the input (using a lightweight module) at each attention layer, or be computed from the first attention layer and then shared among all layers in a stage, or even be computed by an external convolution neural network.  
  % See text for details.
  }
  \label{fig:paca_viz} \vspace{-5mm}
\end{figure}
A picture is worth a thousand words. Seeking solutions that can bridge the semantic gap between those words and raw image data has long been, and remains, a grand challenge in computer vision, machine learning and AI. Deep learning has revolutionized the field of computer vision in the past decade. More recently,  Vision Transformers (ViTs)~\cite{vaswani2017attention,dosovitskiy2020image} have   witnessed remarkable progress in computer vision. ViTs are built on the basis of treating image patches as ``visual tokens" using patch embedding and learning patch-to-patch attention throughout. Unlike the textual tokens that are provided as inputs in natural language processing, visual tokens need to be learned first and continuously refined for more effective learning of ViTs. The patch embedding based tokenizer is a workaround in practice and has a semantic gap with respect to its counterpart, the textual tokenizer.
On one hand, the well-known issue of the quadratic complexity of vanilla Transformer models and the 2D spatial nature of images create a non-trivial task of developing ViTs that are applicable for many vision problems including image classification, object detection and semantic segmentation. On the other hand, explaining trained ViTs requires non-trivial and sophisticated methods~\cite{chefer2021transformer} following the trend of eXplainable AI (XAI)~\cite{gunning2019xai} that has been extensively studied with convolutional neural networks.

\begin{figure}%[t]
  \centering  
  \includegraphics[width=0.48\textwidth]{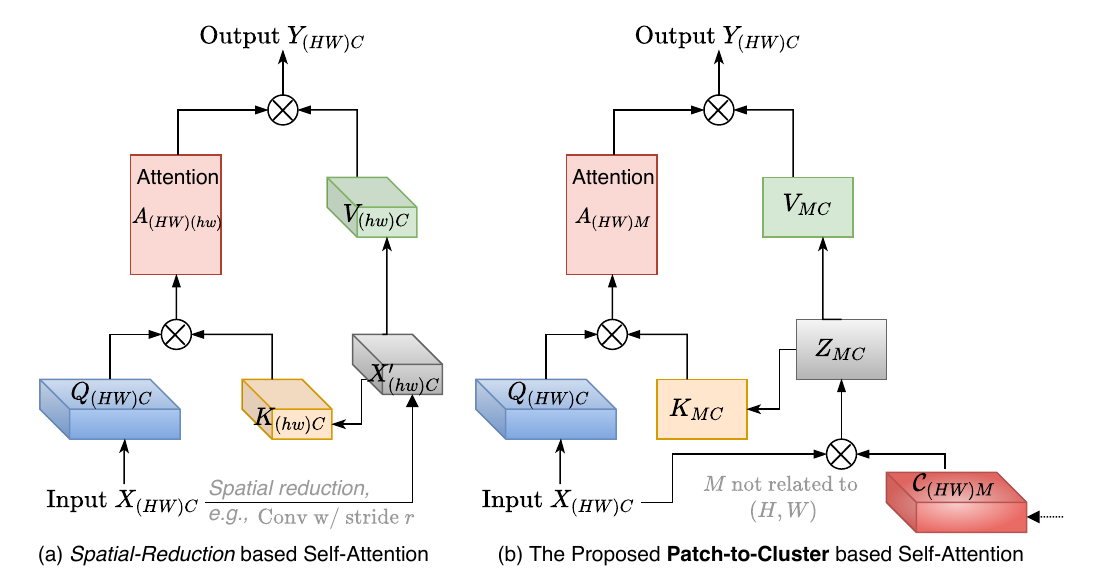}
  \vspace{-4mm}
  \caption{\small Illustration of (a) the spatial reduction based self-attention and (b) the proposed PaCa module in vision applications,  where $(HW)$ represents the number of patches in the input with $H$ and $W$ the height and width respectively, and $M$ a predefined small number of clusters (e.g., $M=100$).  
  % In the proposed PaCa, the $M$ cluster assigment (i.e., $\mathcal{C}_{N, M}$) can either be computed from the input (using a lightweight module, see Eqn.~\ref{eq:clustering}) at each attention layer, or be computed from the first attention layer and then shared among all layers in a stage, or even be computed by an external clustering teacher network (see Fig.~\ref{fig:paca_conv}).
  See text for details.}
  \label{fig:mhsa_viz} \vspace{-4mm}
\end{figure}

To address the quadratic complexity, there have been two main variants developed with great success: One is to exploit the vanilla Transformer model locally using a predefined window size (e.g., $7\times 7$) such as the SWin-Transformer~\cite{liu2021swin}  and the nested variant of ViT~\cite{zhang2021aggregating}. The other is to exploit another patch embedding at a coarser level (i.e., nested patch embedding) to reduce the sequence length (i.e., spatial reduction) before computing the keys and values (while keeping the query length unchanged)~\cite{wang2021pvtv2,wang2021pyramid,wu2021p2t}, as illustrated in Fig.~\ref{fig:paca_viz} (left-bottom) and Fig.~\ref{fig:mhsa_viz} (a). Most of these variants follow the patch-to-patch attention setup used in the vanilla isotropic ViT models~\cite{dosovitskiy2020image}. %Meanwhile, most Transformer  explanation methods are post-hoc and focus on the vanilla ViT models in image classification~\cite{chefer2021transformer}. % Furthermore, explaining ViT models for downstream tasks such as object detection and instance segmentation has not been studied well. 
Although existing ViT variants have shown great results, patch embedding based approaches may not be the best way of learning visual tokens due to the underlying predefined subsampling of the image lattice. Additionally, patch-to-patch attention does not account for the spatial redundancy found in images due to their compositional nature and reusable parts ~\cite{Geman_CompositionSystems}. Thus, it is worth exploring alternative methods towards learning more semantically meaningful visual tokens.  
A question  arises naturally: 
\textit{Can we rethink the patch-to-patch attention mechanism in vision tasks to hit three ``birds" (reducing complexity, facilitating better visual tokenizer and enabling simple forward explainability) with one stone?} 
% The fundamental question underlying the aforementioned challenge is: \textit{What are the visual tokens that should be used for computing keys and values such that Transformer models can be leveraged in an effective and efficient way in computer vision?} Unlike the textual tokens provided as inputs in natural language processing, visual tokens need to be learned first and continuously refined for more effective learning of ViT models.  

\begin{figure*} [t]
    \centering
    \includegraphics[width=0.8\textwidth]{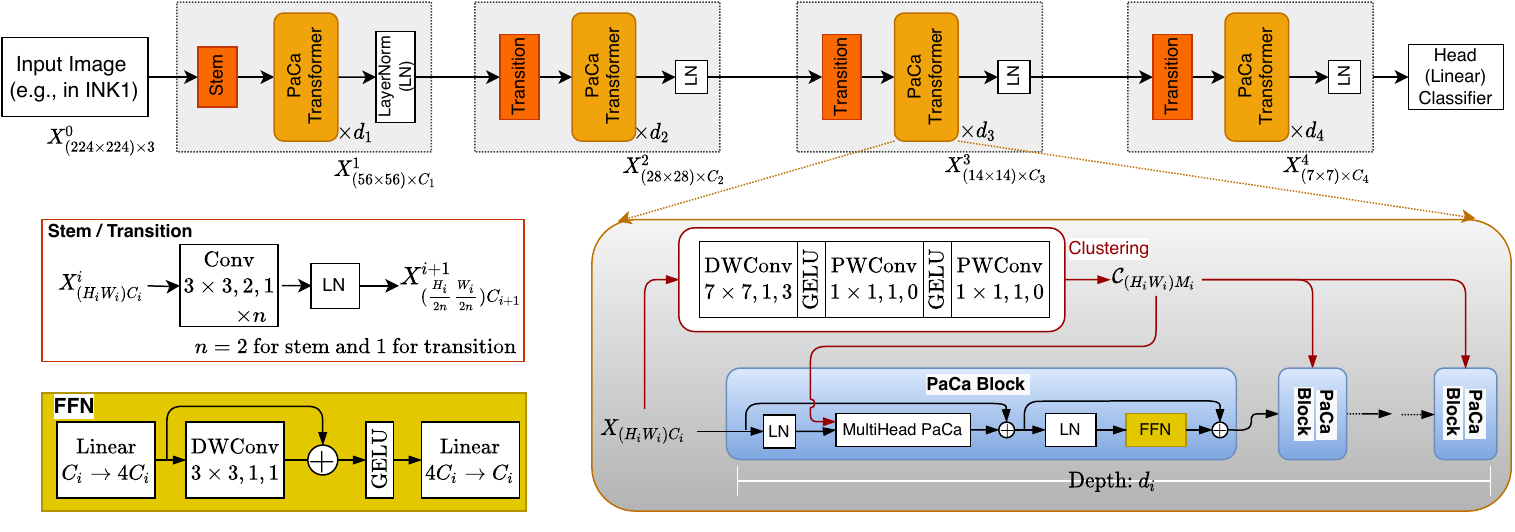}
    \caption{\small Illustration of the proposed PaCa-ViT using the stage-wise convolution-based clustering setting. It consists of four stages each of which has a number, $d_i$ of the proposed PaCa Transformer block.   
    %The proposed PaCa Transformer block exploits the PaCa module in the multi-head self-attention. 
    The {\tt FFN} refers to a feed-forward network.  %The Stem and Transition module are implemented by a convolution module followed by the Layer Normalization (LN)~\cite{ba2016layer}. The downstream task networks are based either on existing work such as the Mask RCNN~\cite{he2017mask} for object detection and instance segmentation, or on the proposed PaCa-based head network for semantic segmentation (to be elaborated in Fig.~\ref{fig:paca_seg_head}).
    See text for details.
    }
    \label{fig:PCAPVT} \vspace{-2mm}
\end{figure*}

As shown in Fig.~\ref{fig:paca_viz} (right) and Fig.~\ref{fig:mhsa_viz} (b), this paper proposes to learn \textbf{Patch-to-Cluster attention (PaCa)}, %under the stage-wise pyramidical architecture (Fig.~\ref{fig:PCAPVT}) of assembling Vision Transformer~\cite{wang2021pyramid,wang2021pvtv2}. 
which provides a straightforward way to address the aforementioned question: 
Given an input sequence $X_{N,C}$ (e.g., $N=H\cdot W$), a  light-weight clustering module finds meaningful clusters by first computing the cluster assignment, $\mathcal{C}_{N, M}$ (Eqn.~\ref{eq:before_clustering} and Eqn.~\ref{eq:clustering}) with a predefined small number of clusters $M$ (e.g., $M=100$). Then, $M$ latent ``visual tokens", $Z_{M,C}$ are formed via simple matrix multiplication between $\mathcal{C}^T_{N, M}$ (transposed) and $X_{N, C}$. In inference, we can directly visualize the clusters $\mathcal{C}_{N,M}$ as heatmaps to reveal what has been captured by the trained models (Fig.~\ref{fig:paca_viz}, right-bottom). The proposed PaCa module induces jointly learning clustering-for-attention and attention-for-clustering in ViT models. We study four aspects of the PaCa module: 
\begin{itemize} [leftmargin=*]
\itemsep0em
\setlength{\parsep}{0pt}
\setlength{\parskip}{0pt}
    \item \textit{Where to compute the cluster assignments?} Consider the stage-wise pyramidical architecture (Fig.~\ref{fig:PCAPVT}) of assembling ViT blocks~\cite{wang2021pyramid,wang2021pvtv2}, a stage consists of a number of blocks. We test two settings: \textit{block-wise} by computing the cluster assignment for each block, or \textit{stage-wise} by computing it only in the first block in a stage and then sharing it with the remaining blocks. Both give comparable performance. The latter is more efficient when the model becomes deeper. 
    \item \textit{How to compute the cluster assignment?} We also test two settings: using 2D convolution or Multi-Layer Perceptron (MLP) based implementation. Both have similar performance. The latter is more generic and sheds light on exploiting PaCa for more general Token-to-Cluster attention (ToCa) in a domain agnostic way. 
    \item \textit{How to leverage an external clustering teacher?} We investigate a method of exploiting a lightweight convolution neural network (Fig.~\ref{fig:paca_conv}) in learning the cluster assignments that are shared by all blocks in a stage. It gives some interesting observations, and potentially pave a way for distilling large foundation models~\cite{bommasani2021opportunities}. 
    \item \textit{What if the number of clusters is known?} We further extend the PaCa module in designing an effective head sub-network for dense prediction tasks such as image semantic segmentation (Fig.~\ref{fig:paca_seg_head}) where the number of clusters $M$ is available based on the ground-truth number of classes and the learned cluster assignment $\mathcal{C}_{N, M}$ has direct supervision. The PaCa segmentation head significantly improves the performance with reduced model complexity. 
\end{itemize}

In experiments, the proposed PaCa-ViT model is tested on the ImageNet-1k~\cite{deng2009imagenet} image classification, the MS-COCO object detection and instance segmentation~\cite{lin2014microsoft} and the MIT-ADE20k semantic segmentation~\cite{zhou2019semantic}. 
It obtains consistently better performance across the three tasks than some strong baseline models including the Swin-Transformers~\cite{liu2021swin} and the PVTv2 models~\cite{wang2021pvtv2}.   
% On the ImageNet-1k validation dataset, the small PaCa-S (22.0M) and base PaCa-B (46.9M)  models achieve top-1 accuracy 83.08\% and 83.96\% respectively. Our PaCa-B outperforms SWin-S~\cite{liu2021swin} (50M) by 0.96\%, SWin-B (88M) by 0.66\%, PVTv2-B3~\cite{wang2021pvtv2} (45.2M) by 0.76\% and PVTv2-B5 (82M) by 0.16\%. On the MS-COCO val2017 dataset, the Mask R-CNN~\cite{he2017mask} model using our PaCa-B as the backbone and trained with 12 epochs (i.e., the 1x schedule) achieves 48.0\% and 42.9\% mean Average Precision (mAP) for bounding boxes and masks respectively, which are better than SWin-B (46.9\% and 42.3\%) and PVTv2-B5 (47.4\% and 42.5\%). On the MIT-ADE20k validation dataset, our segmentor based on the PaCa-B backbone and a PaCa based segmentation head (Fig.~\ref{fig:paca_seg_head}) achieves 50.39\% mIoU (single scale testing), which is significantly better than SWin-B (48.1\% mIoU) and PVTv2-B5 (48.7\% mIoU) with much reduced segmentation head complexity and overall computational cost. 

% \vspace{-2mm}
\section{Related Work and Our Contributions}\label{sec:related}\vspace{-2mm}
Since the pioneering work of ViT~\cite{dosovitskiy2020image}, there has been a vast body of work leveraging and developing variants of the Transformer model~\cite{vaswani2017attention} that dominates the NLP field in computer vision (see a recent survey~\cite{han2020survey}). %We refer to the comprehensive and excellent survey~\cite{han2020survey} for a deeper understanding of the literature. 
We briefly review some of the related efforts on addressing the quadratic complexity of ViT models.  
There has been rapid progress in developing efficient Transformer models. \cite{tay2020efficient} provides an excellent survey of different efforts in the literature. 

One family of approaches is to leverage inductive bias back in the Transformer, including the local window partition based methods~\cite{liu2021swin,child2019generating,parmar2018image,beltagy2020longformer,ainslie2020etc}, random sparse patterns~\cite{zaheer2020big} and the locality-sensitive hashing (LSH) based Reformer~\cite{kitaevreformer}. Although both computational and model performance can be improved, these models achieve them at the expense of limiting the capacity of a self-attention layer due to the locality constraints. Also, sophisticated designs might be needed such as the shifted window and the masked attention method in the SWin-Transformer~\cite{liu2021swin}. The quad-tree based aggregating method proposed in the NesT~\cite{zhang2021aggregating} shows another promising direction. In a similar spirit, the Evo-ViT~\cite{xu2022evo} presents a method of selecting top-$k$ informative tokens for applying the self-attention to reduce the cost. The A-ViT~\cite{yin2022vit} presents a method of halting tokens via reformulating the adaptive computation time (ACT) method. Both Evo-ViT and A-ViT can achieve linear complexity, but they mainly focus on image classification tasks, and it is not clear how to extend them for downstream tasks such as object detection and semantic segmentation. Like the PVT models~\cite{wang2021pyramid,wang2021pvtv2}, our goal is to develop a versatile variant of ViT that can tackle not only image classification, but also many downstream vision tasks which use high-resolution images as inputs and need to retain sufficient high-resolution information throughout.  

Another family of approaches exploits low-rank projections to form a coarser-grained representation of the input sequence, which have shown successful applications for certain NLP tasks such as the LinFormer~\cite{wang2020self}, Nystr\"omformer~\cite{xiong2021nystr} and Synthesizer~\cite{tay2005synthesizer}. Even though these methods retain the capability of enabling each token to attend to the entire input sequence, they suffer from the loss of high-fidelity token-wise information, and on tasks that require fine-grained local information, their performance can fall short of full attention or the aforementioned sparse attention. Similarly, the Performer~\cite{choromanskirethinking} presents a method for Softmax attention kernel approximation via
positive Orthogonal Random features for the Query and the Key.  The proposed PaCa-ViT is motivated by addressing the redundancy of information within patches in patch-to-patch attention in computer vision applications, and shares the spirit of low-rank projection based efficient Transformer models. %Due to the redundancy, the proposed patch-to-cluster attention has the capability of retaining all necessary information in learning. 
Our PaCa is most similar to Linformers~\cite{wang2020self}, but with two main differences: Our PaCa applies clustering (i.e. $\mathcal{C}_{N,M}$) before computing the Key and the Value (Fig.~\ref{fig:mhsa_viz} (b)), unlike Linformers which apply direct projection after the Key and the Value are computed (i.e., $E\cdot K$ and $F\cdot V$, see Eqn.7 in the Linformer paper). Our PaCa reduces the sequence length via a learnable and data-adaptive cluster assignment $\mathcal{C}_{N,M}$, rather than treating the  projection(s), $E$ and $F$, as sequence length specific model parameters.

In addition to achieve the efficiency, driven by XAI~\cite{gunning2019xai} and the natural curiosity of humanity, it is always desirable to understand what is going on inside different ViTs. Most XAI efforts have been focused on convolutional neural networks. More recently, some attention has been attracted to explaining the vanilla isotropic ViT models based on the attention scores themselves. As pointed out in the Improved LRP~\cite{chefer2021transformer}, reducing the explanation to only the attentions scores may be myopic since many other components are ignored. The proposed PaCa provides a direct forward explainer by visualizing the learned cluster assignments as heatmaps, which is complementary to existing approaches.

\textbf{Our Contributions.} This paper makes two main contributions for developing efficient and interpretable variants of Transformers in computer vision applications: (i) It proposes a Patch-to-Cluster Attention (PaCa) module that facilitates learning more expressive and meaningful ``visual tokens" beyond patches in ViTs. It addresses the quadratic complexity of vanilla patch-to-patch attention, while accounting for the spatial redundancy of patches in the patch-to-patch attention.  It also enables a forward explainer for interpreting the trained models based on the learned semantically meaningful clusters. (ii) It proposes a PaCa semantic segmentation head network that is lightweight and more expressive than the widely used UperNet~\cite{xiao2018unified} and the semantic FPN~\cite{kirillov2019panoptic}. With the two main contributions, the proposed PaCa ViTs show superior performance consistently in image classification, object detection and instance segmentation and image semantic segmentation than the prior art including SWin-Transformers~\cite{liu2021swin} and PVTs~\cite{wang2021pyramid,wang2021pvtv2}.  

% \vspace{-2mm}
\section{Approach}\label{sec:approach}\vspace{-2mm}
In this section, we present details of the proposed PaCa module and the resulting PaCa-ViT models.  

\subsection{From Patch-to-Patch Attention to Patch-to-Cluster Attention}\vspace{-2mm}

Denote by $X_{N,C}$ an input sequence consisting of $N$ ``tokens" which are embedded in a $C$-dimensional space. In computer vision, the $N$ tokens are formed via patch embedding. We have $N=H\times W$ where $H$ and $W$ are the height and width of the patch grid respectively~\footnote{We will use the three notations $X_{N,C}, X_{H, W, C}$ and $X_{(HW)C}$ interchangeably in this paper.}. Positional encoding can also be added to counter the permutation insensitivity of the self-attention computation~\cite{vaswani2017attention,dosovitskiy2020image}.

The core of the Transformer model is to compute the scaled dot-product attention in transforming the input $X_{N,C}$ to the output $Y_{N,C}$, 
\begin{align}
     \nonumber A_{N, M} &= \text{Softmax}(\frac{Q_{N,C}\cdot K^T_{M,C}}{\sqrt{C}})_{dim=1},\\  
     Y_{N,C} &= A_{N,M}\cdot V_{M,C}, \label{eq:mhsa}
\end{align}
where $Q_{N,C}, K_{M,C}$ and $V_{M, C}$ are the Query/Key/Value computed from the input $X_{N,C}$, e.g., via linear transformations in the patch-to-patch attention where $M=N$, which leads to the quadratic complexity of computing $A_{N,N}$. The Softmax is applied for each row as indicated by the subscript $dim=1$. In practice, the multi-head self-attention (MHSA) is used to capture the attention in different sub-spaces and fused by a linear projection~\cite{vaswani2017attention}.
To address the quadratic complexity, \textbf{the key} is to ensure $M\ll N$, preferably a predefined constant (e.g., $M=100$) to induce the linear complexity.

To that end, one popular method is to exploit spatial reduction via strided convolution (nested patch embedding) or adaptive average pooling as done in the PVT models~\cite{wang2021pyramid, wang2021pvtv2} (Fig.~\ref{fig:mhsa_viz} (a)). {Note that i) the typically used strided convolution method for spatial reduction does not truly prevent quadratic complexity, but rather reduces it by a ratio corresponding to the patch size, and ii) the adaptive average pooling may suffer from treating each element in a pooling window with equal importance, thus lacking the necessary adaptability and data-driven reweighing capability.}  Meanwhile, the vanilla MLP in the Transformer block has been substituted by the inverted bottleneck block proposed in the MobileNets-v2~\cite{sandler2018mobilenetv2}, termed \textit{MBlock} (the left-bottom of Fig.~\ref{fig:PCAPVT}), which adds a depth-wise convolution in the hidden layer. And, the non-overlapping patch embedding has been replaced by overlapping ones.  With these modifications, positional encoding is not used, partially due to the implicit positional encoding capability of the zero-padding convolutions~\cite{islam2020much} used in the Stem, the Transition modules, and the MBlock. 

On top of the best practices used in PVTv2~\cite{wang2021pvtv2}, we propose the Patch-to-Cluster attention (PaCa), as illustrated in Fig.~\ref{fig:mhsa_viz} (b), which not only achieves the linear complexity (with the overhead of the lightweight clustering module), but also provides a simple cluster assignment visualization method for explaining the attention module. 

\begin{figure*} [t]
    \centering
    \includegraphics[width=0.85\textwidth]{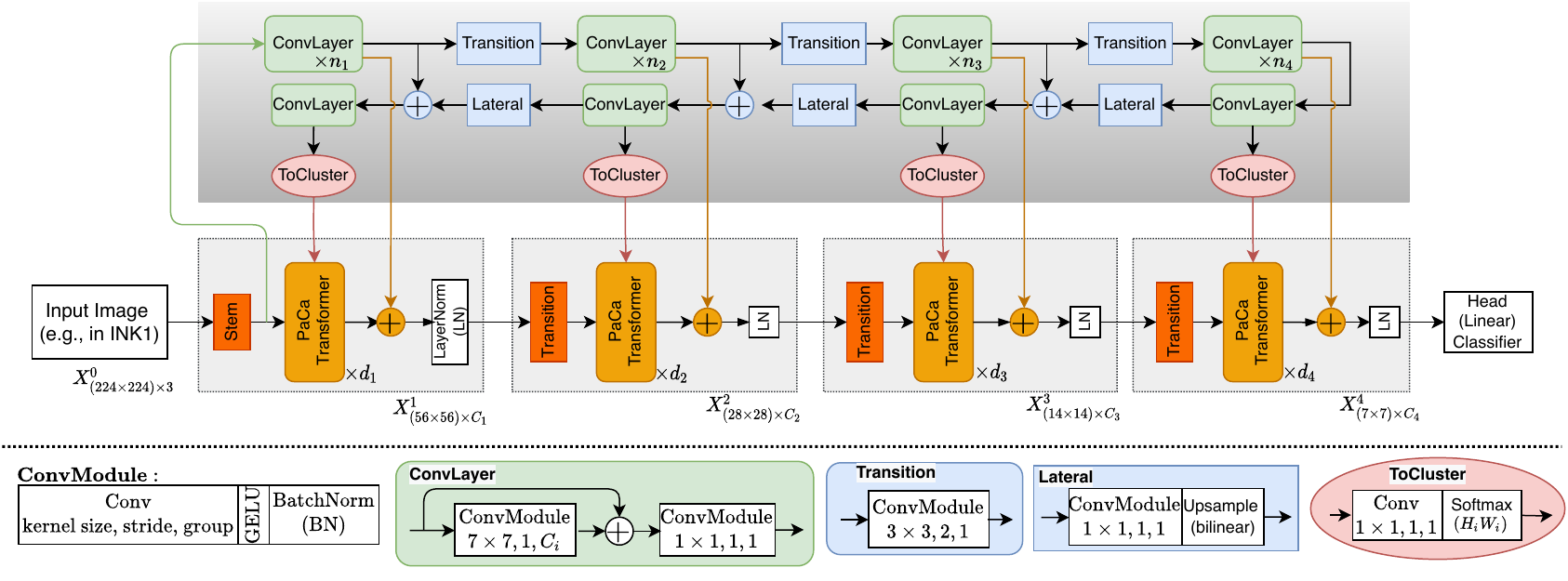}
    \caption{Illustration of computing the cluster assignment via an external clustering teacher network. See text for details.} \vspace{-4mm}
    \label{fig:paca_conv}
\end{figure*}

% \vspace{-2mm}
\subsection{The Proposed Patch-to-Cluster Attention}\vspace{-2mm}
 As shown in Fig.~\ref{fig:mhsa_viz} (b), given an input sequence $X_{N, C}$ and a predefined number $M$ of clusters (e.g., $M=100$), we first compute the cluster assignment $\mathcal{C}_{N, M}$, whose goal is to cluster the input sequence into $M$ latent ``visual tokens", 
 \begin{equation}
    Z_{M,C} = \text{LayerNorm}(\mathcal{C}^T_{N,M}\cdot X_{N,C}), \label{eq:pca}
\end{equation}
which then is used to compute the Key, $K_{M,C}$ and the Value, $V_{M,C}$ via linear transformations in computing the attention (Eqn.~\ref{eq:mhsa}).  
We present two methods of computing the cluster assignment $\mathcal{C}_{N, M}$:  One is the \textbf{onsite clustering} as illustrated in Fig.~\ref{fig:PCAPVT}, and the other is the \textbf{external clustering} via a lightweight teacher network as illustrated in Fig.~\ref{fig:paca_conv}. 

\vspace{-2mm}
\subsubsection{Onsite Clustering} \vspace{-2mm} 
For the \textit{onsite clustering} method (Fig.~\ref{fig:PCAPVT}), we have,
\begin{align}    
    \mathcal{C}_{N, M} &= \text{Clustering}(X_{N,C};\theta), 
\end{align}
where $\theta$ collects the parameters of the clustering module. We investigate two simple designs in this paper: 

\textbf{i) Clustering via Convolution}: It uses depth-wise convolution (DWConv) and point-wise convolution (PWConv), 
\begin{align}
    X_{N,C}&\xrightarrow[k=7,s=1]{\text{DWConv+GELU}}\cdot \xrightarrow[k=1,s=1]{\text{PWConv+GELU}} U_{N,C}, \label{eq:before_clustering}\\
    U_{N,C}&\xrightarrow[k=1,s=1]{\text{PWConv}}\cdot\xrightarrow[\text{along }N]{\text{Softmax}} \mathcal{C}_{N, M}, \label{eq:clustering}
\end{align}
where the first DWConv module uses a relatively large $k=7$ kernel with stride $s=1$ and zero padding $3$, which is to integrate local information with a larger receptive field.

\textbf{ii) Clustering via MLP}: To be more generic by eliminating the dependence on 2D convolution in Eqn.~\ref{eq:before_clustering}, it utilizes a MLP implementation,
\begin{equation}
    X_{N,C}\xrightarrow[C\rightarrow 4C]{\text{Linear+GELU}}\cdot\xrightarrow[4C\rightarrow M]{\text{Linear}}\cdot\xrightarrow[\text{along }N]{\text{Softmax}} \mathcal{C}_{N, M},\label{eq:before_clustering_1}
\end{equation}
where the expansion ratio of the hidden layer of the MLP is set to 4 by default.

To encourage forming meaningful clusters (i.e., visual tokens) that can capture underlying visual patterns that are often spatially sparse, we apply Softmax along the spatial dimension in  Eqns.~\ref{eq:clustering} and~\ref{eq:before_clustering_1}, which also enables directly visualizing $ \mathcal{C}_{N, M} $ as $M$ heatmaps \textit{for diagnosing the interpretability} of a trained model at the instance level in a forward computation.

\textbf{Where to compute $\mathcal{C}_{N,M}$ in the onsite clustering setting?} As aforementioned, $\mathcal{C}_{N,M}$ can be computed in either a block-wise or a stage-wise setting (the right-bottom of Fig.~\ref{fig:PCAPVT}). We observed that the latter is not only more computationally efficient, but also more effective in terms of accuracy in our ablation study. Intuitively, sharing the cluster assignment in a stage facilitates consistency between different Transformer blocks, and may induce meaningful latent features at the front-end of a stage (Eqn.~\ref{eq:before_clustering}) based on the collective feedback from all the blocks in a stage during training. 

\vspace{-2mm}
\subsubsection{Understanding the PaCa Module} \vspace{-2mm}
Intuitively, computing $Z_{M,C}$ via the matrix multiplication (Fig.~\ref{fig:paca_viz} (b) and Eqn.~\ref{eq:pca}) can be understood as a depth-wise global weighted pooling of the input $X_{N,C}$ with learned weights, $\mathcal{C}_{N,M}$. 
It can also be seen as a dynamic MLP-Mixer~\cite{tolstikhin2021mlp} with data-driven weight parameters $\mathcal{C}_{N,M}$ for the spatial transformation and integration component, rather than using top-down model parameters, making it more flexible by not restricting the trained models to a specific input size. The learned clusters (latent visual tokens) $Z_{M,C}$ share similar spirit to the class-token(s) or task prompts used in the vanilla ViT models (single class token) and its variants with multiple class-tokens. The former are data driven, while the latter are treated as model parameters. 

Furthermore, the learned clustering assignment $\mathcal{C}_{N, M}$ has the same form of the attention matrix $A_{N, M}$ (Eqn.~\ref{eq:mhsa}). Computing $Z_{M,C}$ (Eqn.~\ref{eq:pca}) can thus be understood as performing cross-covariance attention (XCA)~\cite{el2021xcit}. 

Learning $\mathcal{C}_{N, M}$  itself can be understood as a way of learning better visual tokens to bridge the gap between patches and the textual tokens used in natural language processing Transformer models. This type of visual tokenizer has also been observed to be useful in integrating Transformer models on top of convolution neural networks (CNNs) such as ResNets in Visual Transformer~\cite{wu2020visual}.    %In line of this, the proposed PCA module has a connection with the matrix decomposition method~\cite{} that is recently proposed as an alternative to self-attention.  

\vspace{-2mm}
\subsubsection{External Clustering} \vspace{-2mm}
With the onsite clustering setting, the cluster assignments $\mathcal{C}_{N, M}$'s at the early stages are based on the low-to-middle level information. To address this issue, we are inspired by three lines of work: the feature pyramid network (FPN)~\cite{lin2017feature} that is widely used for integrating visual information at different levels,  the slow-fast thinking paradigm~\cite{kahneman2011thinking} (i.e., System 1 v.s. System 2) recently prompted for inducing reasoning capabilities in neural networks~\cite{goyal2022inductive}, and the empirical observations of Transformers focusing more on low-frequency information and CNNs focusing more on high-frequency information~\cite{park2022vision}. We introduce an \textit{external clustering} teacher CNN (Fig.~\ref{fig:paca_conv}) that is concurrently trained with the PaCa ViT end-to-end. 
To be lightweight, we use the ConvMixer layer~\cite{trockman2022patches} in the FPN-style CNN clustering teacher network~\footnote{Many other lightweight CNNs, such as the MobileNets~\cite{sandler2018mobilenetv2}, can be straightforwardly applied.}.  

With the clustering teacher network, we first compute all the stage-wise clustering assignments, and then learn the PaCa ViT. We also integrate the stage outputs from the teacher network into the PaCa ViT. 
The clustering teacher network can be interpreted as a fast learner to provide informative guidance (the cluster assignment) to the relatively slower PaCa learner. It can also be intuitively interpreted as a type of working memory~\cite{christophel2017distributed} that ``manipulate" the input data to facilitate the ``post-processing" via the PaCa. In addition, this integration may facilitate harnessing the joint expressive power of learning high-frequency information  by the CNN teacher and of learning low-frequency information by the Transformer based models~\cite{park2022vision}. 

% \textit{Remarks}: The proposed CNN and Transformer integration bridged by the clustering module potentially provides a more comprehensive understanding of diverse image data in continual learning with streaming tasks~\cite{mccloskey1989catastrophic,thrun1995lifelong}. We also note that the lightweight CNN teacher could be replaced by pretrained image encoder such as the CLIP models~\cite{radford2021learning}, which may provide alternative ways other than prompt-based finetuning strategies~\cite{liu2021pre} for leveraging the pretrained large foundation models (LFMs)~\cite{bommasani2021opportunities} with potentially interesting results in different applications. We can further design progressive training methods to distill the LFM-based clustering teacher networks into the onsite clustering modules such that the teacher network will not be needed in inference, similar in spirit to the contrastive teacher-student self-supervised learning such as the MoCo models~\cite{he2020momentum}. %This is consistent to the aforementioned exploration of exploiting self-supervised learning loss functions in regularizing the diversity and meaningfulness of the clusters, with the pretrained LFMs as the loss functions. 

\vspace{-2mm}
\subsubsection{Task-Specific Tuning of $M$} \vspace{-2mm}
The number of clusters $M$ can be changed accounting for the task specific information. For example, when using an ImageNet-1k pretrained PaCa model that uses a relative small $M$ (e.g., 100) as the backbone in a downstream task (e.g., the 150-class MIT-ADE20k dataset), we observe that we can change $M$ to a large number (e.g., 200), which only results in minor changes of the network (e.g., the PWConv in Eqn.~\ref{eq:clustering}) and has no training issue observed in our experiments (see Sec.~\ref{sec:segmentation}).

\vspace{-2mm}
\subsubsection{Complexity Analyses} \vspace{-2mm}
Compared with PVTv2~\cite{wang2021pvtv2}, our PaCa (Eqn.~\ref{eq:pca}) leads to \textit{linear} complexity in computing the self-attention matrix (Eqn.~\ref{eq:mhsa}) since the number of clusters $M$ is predefined and fixed in our PaCa-ViT models. This advantage is achieved at the expense of the overhead cost in Eqn.~\ref{eq:before_clustering} or Eqn.~\ref{eq:clustering} and the matrix multiplication in Eqn.~\ref{eq:pca}. For relatively small images (e.g., in image classification), the overhead cost slightly outweighs the reduction in  computing the self-attention matrix (see Sec.~\ref{sec:classification}). For large images (e.g., in object detection and instance semantic segmentation), the overhead cost is well paid off, leading to significant reduction of computing cost and memory footprint (see Sec.~\ref{sec:detection} and Sec.~\ref{sec:segmentation}).

\vspace{-1mm}
\subsection{Network Interpretability via  PaCa }\label{sec:interpretability} \vspace{-2mm}
To select the most important clusters in $\mathcal{C}_{N, M}$  for an input image $I$ in a vision task (e.g., image classification), we adopt a straightforward approach. We use the clustering assignment maps $\mathcal{C}_{N, M}$ before the Softmax and then apply the Sigmoid transformation. For each cluster $m$, we reshape the slice $\mathcal{C}_{N, m}$ back to a 2D spatial heatmap , denoted by $\mathcal{H}^m_{h, w}$.  We first compute a binary mask by keeping locations whose clustering scores are greater than the mean score, 
\begin{equation}
    \mathbb{M}^m_{h, w} = \mathcal{H}^m_{h, w} > mean(\mathcal{H}^m_{h, w}),
\end{equation}
which is then upsampled to the resolution of input images (e.g., 224$\times$ 224) using the nearest interpolation, denoted by $\mathbb{M}^m$.  The upsampled mask is used to mask the input image $I$ that can be correctly classified by the model, and we have, 
\begin{align}
I^{m} = I \odot \mathbb{M}^m,
\end{align}
where $\odot$ represents element-wise product. $I^{m}$ is then used as the input to the model. 

We divide the learned clusters into two groups: the positive group in which a masked image $I^m$ can still predict the ground-truth label, and the negative group in which a masked image $I^{m'}$ has the wrong predicted label. The positive group means that the masked portion in an image based on the clustering assignment retains sufficient information.

\begin{figure}
    \centering
    \includegraphics[width=0.45\textwidth]{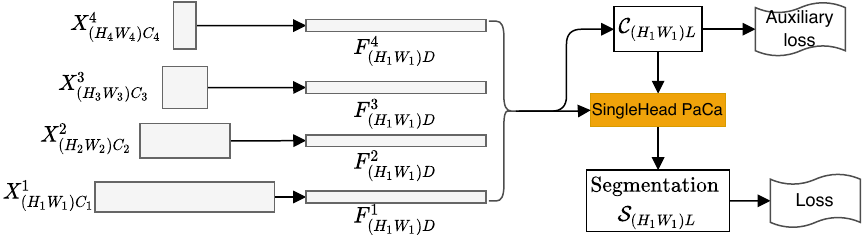}
    \caption{\small The proposed PaCa head network for semantic segmentation. The feature pyramid from the backbone (Fig.~\ref{fig:PCAPVT}) is projected to a $D$-dim feature space and resized to the resolution of the first feature layer (via bilinear interpolation). $L$ is the number of classes (e.g., 150 in the MIT-ADE20k dataset). Both the losses are cross-entropy, but with a smaller weight for the auxiliary loss.}
    \label{fig:paca_seg_head} \vspace{-3mm}
\end{figure}

\subsection{The PaCa Segmentation Head}\label{sec:paca_seg_head}\vspace{-2mm}
The image semantic segmentation task can provide direct supervision signals to the clustering assignment $\mathcal{C}_{N,M}$ (with $M=L$ the number of the ground-truth classes). We elaborate the design shown in Fig.~\ref{fig:paca_seg_head} in this section and present results in Sec.~\ref{sec:segmentation}. With the feature pyramid $X^i_{(H_iW_i)C_i}$'s (e.g., $i=1,2,3,4$) from the backbone, we first transform each pyramid layer into a $D$-dim feature space, $F^i_{H_1W_1}D$ via a vanilla convolution block (ConvBlock) consisting of $1\times 1$ convolution, BatchNorm~\cite{BatchNorm} and ReLU~\cite{AlexNet}, followed by a bilinear upsampling (for layers other than the first one). Denote by $F_{(H_1W_1)4D}$ as the concatenated feature map as the multi-scale fused input, from which the clustering assignment $\mathcal{C}_{N_1,L}$ ($N_1=H_1\times W_1$) is computed,
\begin{equation}
    F_{N_1,4D}\xrightarrow[4D\rightarrow D]{\text{ConvBlock}} \cdot \xrightarrow[k=1,s=1]{\text{PWConv}}\cdot\xrightarrow[\text{along }N_1]{\text{Softmax}} \mathcal{C}_{N_1,L}.
\end{equation}

Then, a single-head PaCa module is used with the Query, Key and Value computed as follows,
\begin{align}
    \text{Query: } & F_{N_1,4D} \xrightarrow[]{\text{ConvBlock}}  Q_{N_1,D}, \\
    \text{Clusters: } & Z_{L,4D} = \mathcal{C}_{N_1,L}^T \cdot F_{N_1,4D}, \\
    \text{Key \& Value: } & Z_{L,4D} \xrightarrow[]{\text{LinearBlock}} K_{L,D}, V_{L,D}, 
    % \text{Value: } & Z_{L,4D} \xrightarrow[]{\text{LinearBlock}} V_{L,D}, 
\end{align}
where a LinearBlock consisting of a linear projection layer, a BatchNorm1D and the ReLU. The output of the PaCa module is computed by,
\begin{equation}
    \text{Softmax}_{L}(Q_{N_1,D}\cdot K_{L,D}^T)\cdot V_{L,D}\xrightarrow[]{\text{ConvBlock}} \mathbf{F}_{N_1,D}.
\end{equation}
The final segmentation result is regressed via,
\begin{equation}
    \mathcal{S}_{N_1, L} = \text{PWConv}(\mathbf{F}_{N_1,D}).
\end{equation}

\vspace{-2mm}
\section{Experiment}\label{sec:experiment} \vspace{-2mm}
In this section, we present experimental results of the proposed method in ImageNet-1k (IN1K)~\cite{deng2009imagenet} classification, MS-COCO 2017 object detection and instance segmentation~\cite{lin2014microsoft} and MIT-ADE20k semantic segmentation~\cite{zhou2019semantic}. In implementation, we use the popular {\tt timm} PyTorch toolkit~\cite{rw2019timm} for image classification, the {\tt mmdetection} and {\tt mmsegmentation} toolkits\cite{chen2019mmdetection} for object detection and semantic segmentation respectively. 

% Due to the limited computing resources, 
We mainly test three stage-wise onsite clustering-via-convolution PaCa models (Fig.~\ref{fig:PCAPVT}): PaCa-Tiny (12.2M), PaCa-Small (22.0M) and PaCa-Base (46.9M). For comparisons, we one stage-wise onsite clustering-via-MLP PaCa mode: PaCa$^{mlp}$-Small (22.6M). For onsite clustering, the PaCa is used in the first three stages with the number of clusters $M=100$. We also test two stage-wise external clustering based PaCa models (Fig.~\ref{fig:paca_conv}): PaCa$^{ec}$-small (21.1M) and PaCa$^{ec}$-base (47.65M), both with $M=100$ for all the four stages. Detailed architectural specifications are provided in the supplementary. The training recipes are the same with the prior art and provided in the supplementary too.

% whose architectural specifications are provided in the supplementary, and clustering assignment modules use $M=100$ clusters and are shared within each stage (see comparisons in the ablation study in Sec.~\ref{sec:ablation}). The training receipts are the same with the prior arts and are provided in the supplementary materials. Due to space limit, \textbf{all the qualitative results on visualizing the learned clusters are also provided in the supplementary materials.}

\subsection{Image Classification} \label{sec:classification}\vspace{-2mm}

\begin{table}
    \centering
    \resizebox{0.43\textwidth}{!}{
    \begin{tabular}{l|c|c|l}
         Method & \#Params (M)$\downarrow$ & FLOPs (G) $\downarrow$ & Top-1 Acc. (\%)$\uparrow$ \\ \toprule 
         % ResNet-18~\cite{he2016deep} & 11.7 & 1.8 &  69.8 \\ 
         DeiT-T/16~\cite{touvron2021training} & 5.7M & 1.3 & 72.2 \\
         PVT-T~\cite{wang2021pyramid} & 13.2 & 1.9 & 75.1 \\
         PVTv2-B1~\cite{wang2021pvtv2} & 14.0 & \underline{2.1} & \underline{78.7} \\ 
         \textbf{PaCa-Tiny (ours)} & 12.2 & 3.2  &  \textbf{80.9} \textcolor{blue}{$\uparrow$2.2} \\ \midrule
         % ResNet-50~\cite{he2016deep} & 25.6 & 4.4 & 76.1\\
        % ResNeXt-50-32x4d~\cite{xie2017aggregated} & 25.0 & 4.3 &77.6\\
        % RegNetY-4G~\cite{radosavovic2020designing} & 21.0 & 4.0 & 80.0\\
        DeiT-S/16~\cite{touvron2021training} & 22.1 & 4.6 &79.9\\
        T2T-ViT$_t$-14~\cite{yuan2021tokens} & 22.0 & 6.1 & 80.7\\
        PVT-S~\cite{wang2021pyramid} & 24.5 & 3.8 &79.8\\
        TNT-S~\cite{han2021transformer} & 23.8 & 5.2 &81.3\\
        SWin-T~\cite{liu2021swin} & 29.0 & 4.5 &81.3\\
        CvT-13~\cite{wu2021cvt} & 20.0 & 4.5 &81.6\\
        % CoaT-Lite-S~\cite{xu2021co} &20.0 & 4.0 &81.9\\
        Twins-SVT-S~\cite{chu2021twins} & 24.0 & 2.8 & 81.3 \\
        FocalAtt-Tiny~\cite{yang2021focal} & 28.9 & 4.9 & 82.2 \\
         PVTv2-B2~\cite{wang2021pvtv2} & 25.4 & 3.9 & 82.0 \\ 
          PVTv2-B2-li~\cite{wang2021pvtv2} & 22.6 & \underline{4.0} & \underline{82.1} \\ 
         {PaCa-Small (ours)} & 22.0 &  5.5 &  {83.08} \textcolor{blue}{$\uparrow$0.98} \\
         {PaCa$^{mlp}$-Small (ours)} & 22.6 &  5.9 &  {83.13} \textcolor{blue}{$\uparrow$1.03} \\
         \textbf{PaCa$^{ec}$-Small (ours)} & 21.1 &  5.4 &  \textbf{83.17} \textcolor{blue}{$\uparrow$1.07} \\ \midrule
         % ResNet101~\cite{he2016deep} & 44.7 & 7.9 & 77.4 \\
         % ResNeXt101-32x4d~\cite{xie2017aggregated} & 44.2 & 8.0 & 78.8 \\
         % ResNeXt101-64x4d~\cite{xie2017aggregated} &  83.5 & 15.6 & 79.6 \\
        % RegNetY-8G~\cite{radosavovic2020designing} & 39.0 & 8.0 & 81.7 \\
        T2T-ViT$_t$-19~\cite{yuan2021tokens} & 39.0 & 9.8 & 81.4\\
        T2T-ViT$_t$-24~\cite{yuan2021tokens} & 64.0 & 15.0 & 82.2\\
        PVT-M~\cite{wang2021pyramid} & 44.2 & 6.7 & 81.2\\
        PVT-L~\cite{wang2021pyramid} & 61.4 & 9.8 & 81.7 \\
        CvT-21~\cite{wu2021cvt} & 32.0 & 7.1 &82.5\\
        TNT-B~\cite{han2021transformer} & 66.0 & 14.1 & 82.8 \\
        SWin-S~\cite{liu2021swin} & 50.0 & 8.7 & 83.0 \\
        SWin-B~\cite{liu2021swin} & 88.0 & 15.4 & 83.3 \\
        Twins-SVT-B~\cite{chu2021twins} & 56.0 & 8.3 & 83.2 \\   
        Twins-SVT-L~\cite{chu2021twins} & 99.2 & 14.8 & 83.7 \\
        % MLP-Mixer-B/16~\cite{tolstikhin2021mlp} & 59.9 & 12.7 & 76.4 \\
        % gMLP-B~\cite{liu2021pay} & 73.4 & 15.8 & 81.6 \\        
        % ResMLP-B24~\cite{touvron2022resmlp} & 129.1 & 23.0 & 81.0 \\ 
        % PoolFormer-m36~\cite{yu2022metaformer} & 56.2 & 8.8 &  82.1 \\
        % PoolFormer-m48~\cite{yu2022metaformer} & 73.5 & 11.6  & 82.5 \\
        FocalAtt-Small~\cite{yang2021focal} & 51.1 & 9.4 & 83.5 \\
        FocalAtt-Base~\cite{yang2021focal} & 89.8 & 16.4 & 83.8 \\
        PVTv2-B3~\cite{wang2021pvtv2} & 45.2 & \underline{6.9} & \underline{83.2}\\
        PVTv2-B4~\cite{wang2021pvtv2} & 62.6 & 10.1 & 83.6 \\
        PVTv2-B5~\cite{wang2021pvtv2} & 82.0 &  11.8 & 83.8 \\
        {PaCa-Base (ours)} & 46.9 &  9.5 &  {83.96} \textcolor{blue}{$\uparrow$0.76}  \\
        \textbf{PaCa$^{ec}$-Base (ours)} & 46.7 &  9.7 &  \textbf{84.22} \textcolor{blue}{$\uparrow$1.02}  \\ \bottomrule         
    \end{tabular}
    }
    \caption{Top-1 accuracy comparison in IN1K validation set using the single center crop ($224\times 224$) evaluation protocol. The relative  improvement of our PaCa models are computed with respect to the PVTv2 models (underlined) with similar parameters. } \vspace{-4mm}
    \label{tab:in1k} 
\end{table}

The IN1K classification dataset~\cite{deng2009imagenet} consists of about $1.28$ million images for training, and $50,000$ for validation, from $1,000$ classes. All models are trained on the training set
for fair comparisons and report the Top-1 accuracy on the validation set. We follow the training recipe used by the PVTv2 which in turn is adopted from the DeiT~\cite{touvron2021training}. %Detail settings will be provided in the supplementary material. 

\textbf{Accuracy.} Table~\ref{tab:in1k} shows the comparisons. The proposed PaCa ViT obtains consistently better performance than many variants of ViTs including the baseline PVTv2, which justifies the effectiveness of the proposed patch-to-cluster attention. With the onsite stage-wise clustering setting, clustering-via-MLP (Eqn.~\ref{eq:before_clustering_1}) is slightly better than clustering-via-convolution (Eqns.~\ref{eq:before_clustering} and~\ref{eq:clustering}). The external clustering (Fig.~\ref{fig:paca_conv}) outperforms the onsite clustering (Fig.~\ref{fig:PCAPVT}) slightly.  Fig.~\ref{fig:paca_xai} shows some examples of the learned clusters. 
\textbf{Efficiency.} In terms of efficiency based on FLOPs, the proposed PaCa models are slightly worse at the resolution of $224\times 224$ in IN1K. As aforementioned, the efficiency will significantly improve and outperform other variants in downstream tasks with higher resolution images. 

% \textbf{Interpretability.}  We show examples of directly visualizing the learned clustering heatmaps in the supplementary document, which are often meaningful and shed light on pursuing more expressive clustering modules in implementation (see remarks and discussions therein).  

\begin{table}[t]
    \centering
    \resizebox{0.49\textwidth}{!}{
    \begin{tabular}{l|l|l|ccc|ccc}

        Backbone & \#Params (M) & FLOPs (G)  & AP$^b$ & AP$^b_{50}$ & AP$^b_{75}$ & AP$^m$ & AP$^m_{50}$ & AP$^m_{75}$ \\ \toprule        
         % ResNet-18~\cite{he2016deep} & 31.2 & -  & 36.9 &57.1 &40.0 &33.6& 53.9& 35.7 \\
            PVT-T~\cite{wang2021pyramid} & 32.9 & - & 39.8& 62.2 &43.0& 37.4& 59.3& 39.9 \\
            PVTv2-B1~\cite{wang2021pvtv2} & 33.7& 259$^*$ &  {41.8}& 64.3& 45.9 &{38.8} &61.2& 41.6 \\ 
            \textbf{PaCa-Tiny (ours)} & 32.0& 252$^*$  & \textbf{43.3}& 66.0& 47.5 &\textbf{39.6} &62.9& 42.4 \\ \midrule 
            ResNet-50~\cite{he2016deep}  & 44.2& 260  & 41.0& 61.7& 44.9 &37.1 &58.4 &40.1 \\
            SWin-T~\cite{liu2021swin} & 47.8 & 264 & 43.7 & 66.6 &47.7& 39.8 &63.3 &42.7\\
            Twins-SVT-S~\cite{chu2021twins} & 44.0 &228 &43.4& 66.0 &47.3 &40.3 &63.2 &43.4 \\
            FocalAtt-T~\cite{yang2021focal} & 48.8 &291 &44.8& 67.7& 49.2 &41.0 &64.7 &44.2 \\
            PVT-S~\cite{wang2021pyramid} & 44.1& 245 & 43.0 &65.3& 46.9 &39.9& 62.5& 42.8 \\
            % PVTv2-b2-li~\cite{wang2021pvtv2}& 42.2 & 246.0 &44.1 &66.3& 48.4 &40.5& 63.2 &43.6 \\
            PVTv2-B2~\cite{wang2021pvtv2}&  45.0& 325$^*$  & {45.3}& 67.1& 49.6 &{41.2} &64.2 &44.4 \\ 
            {PaCa-Small (ours)}&  41.8& 296$^*$ & {46.4}& 68.7 & {50.9} &{41.8} &{65.5} &{45.0} \\
            \textbf{PaCa$^{mlp}$-Small (ours)}&  42.4& 303$^*$ & \textbf{46.6}& 69.0 & {51.3} &\textbf{41.9} &{65.7} &{45.0} \\ 
            {PaCa$^{ec}$-Small (ours)}&  40.9& 292$^*$ & {45.8}& 68.0 & {50.3} &{41.4} &{64.9} &{44.5} \\ \midrule 
            SWin-S~\cite{liu2021swin} & 69.1 & 354 & 46.5 & 68.7 &51.3 &42.1 &65.8 &45.2 \\
            SWin-B~\cite{liu2021swin} &107.1 &497 &46.9 &69.2 &51.6 &42.3 &66.0 &45.5\\
            FocalAtt-S~\cite{yang2021focal} & 71.2 &401 &47.4 &69.8 &51.9 &42.8 &66.6 &46.1 \\
            FocalAtt-B~\cite{yang2021focal} & 110.0 &533 &47.8& 70.2& 52.5& {43.2} &67.3 &46.5\\
            Twins-SVT-B~\cite{chu2021twins} & 76.3 & 340 & 45.2 & 67.6 & 49.3 & 41.5 & 64.5 & 44.8 \\
            PVT-M~\cite{wang2021pyramid} & 63.9 & 302 & 42.0 & 64.4 & 45.6 & 39.0 & 61.6  & 42.1 \\
            PVT-L~\cite{wang2021pyramid} & 81.0 & 364 & 42.9 & 65.0 & 46.6 & 39.5 & 61.9 & 42.5 \\
            PVTv2-B3~\cite{wang2021pvtv2}& 64.9 & 413$^*$ & 47.0  & 68.1 &  51.7 &  42.5 &  65.7 &  45.7 \\
            PVTv2-B4~\cite{wang2021pvtv2}& 82.2 & 516$^*$ & 47.5 &68.7& 52.0& 42.7& 66.1 &46.1 \\
            PVTv2-B5~\cite{wang2021pvtv2}& 101.6 & 573$^*$ & 47.4 & 68.6 & 51.9& 42.5 &65.7 & 46.0 \\
            {PaCa-Base (ours)}&  66.6& 373$^*$ & {48.0}& 69.7 & 52.1 & {42.9} &{66.6} &{45.6} \\ 
            \textbf{PaCa$^{ec}$-Base (ours)}&  61.4 & 372$^*$ & \textbf{48.3}& 70.5 & 52.6 & \textbf{43.3} &{67.2} &{46.6} \\\bottomrule 
    \end{tabular}
    } \vspace{0.2em}
    \caption{Object detection and instance segmentation on MS-COCO val2017~\cite{lin2014microsoft} using the IN1K pretrained backbones and the Mask R-CNN~\cite{he2017mask} with the 1x (12-epoch) training schedule in training. FLOPs are computed at the input resolution of $1280\times 800$. $^*$computed using the  \href{https://github.com/zhijian-liu/torchprofile}{torchprofile} package.}
    \label{tab:coco} \vspace{-1mm}
\end{table}

\begin{table}[h]
    \centering
    \resizebox{0.48\textwidth}{!}{
    \begin{tabular}{l|  l|c|c| l }
        
        Backbone & Head & \#Params (M) & FLOPs (G)  & mIOU \\ \toprule 
        PVT-T~\cite{wang2021pyramid} & \multirow{9}{*}{\parbox{2cm}{\centering{Semantic  FPN~\cite{kirillov2019panoptic}}}}  & 17.0 & 33.2 & 35.7 \\ \cline{3-5}
        PVT-S & & 28.2 & 44.5 & 39.8\\ \cline{3-5}
        PVT-M & & 48.0 & 61.0 & 41.6 \\ \cline{3-5}
        PVT-L & & 65.1 &  79.6 & 42.1 \\ \cline{3-5}
        PVTv2-B1~\cite{wang2021pvtv2} & & 17.8 & 34.2 & 42.5  \\ \cline{3-5}
        PVTv2-B2 &  & 29.1 & 45.8 & 45.2   \\ \cline{3-5}        
        PVTv2-B3 &  & 49.0 & 62.4  & 47.3   \\ \cline{3-5}
        PVTv2-B4 &  & 66.3 & 81.3 & 47.9   \\ \cline{3-5}
        PVTv2-B5 &  & 85.7 & 91.1 & 48.7   \\ \toprule
        SWin-T~\cite{liu2021swin} & \multirow{3}{*}{UperNet~\cite{xiao2018unified}} & 60 & 941 & 44.5 \\ \cline{3-5}
        SWin-S & & 81 & 1038 & 47.6 \\ \cline{3-5}
        SWin-B & & 121 &  1188 &  48.1 \\ \toprule
        FocalAtt-T~\cite{yang2021focal} & \multirow{3}{*}{UperNet~\cite{xiao2018unified}} & 62 & 998 & 45.8\\ \cline{3-5}
        FocalAtt-S & & 85 & 1130 & 48.0 \\ \cline{3-5}
        FocalAtt-B & & 126 & 1354 & 49.0 \\ \toprule
        PaCa-Tiny (ours) & \multirow{3}{*}{UperNet~\cite{xiao2018unified}} & 41.6 & 229.9$^*$ & 44.49    \\ \cline{3-5}
        PaCa-Small  (ours) &  & 51.4 & 242.7$^*$  & 47.6\\ \cline{3-5}
        PaCa-Base  (ours) &  & 77.2 & 264.1$^*$  & 49.67  \\ \midrule
        PaCa-Tiny (ours) &  \multirow{6}{*}{PaCa  (ours)} & 13.3 & 34.4$^*$  & 45.65  \\ \cline{3-5}    
        PaCa-Small  (ours) &  & 23.2 & 47.2$^*$  & 48.3   \\ \cline{3-5}    PaCa$^{mlp}$-Small  (ours) &  & 24.0 & 50.0$^*$  & 48.2   \\ \cline{3-5}  
        PaCa$^{ec}$-Small  (ours) &  & 22.2 & 46.4$^*$  & 46.2   \\ \cline{3-5}
        \textbf{PaCa-Base (ours)} &  & 48.0 & 68.5$^*$ &  \textbf{50.39} \\ \cline{3-5}
        PaCa$^{ec}$-Base  (ours) &  & 48.8 & 68.7$^*$ & 48.4   \\ 
        \bottomrule
    \end{tabular}}\vspace{0.2em}
    \caption{Semantic segmentation on MIT-ADE20k~\cite{zhou2019semantic} with the crop size $512\times 512$ using the IN1K pretrained backbones. FLOPs are computed at the input resolution of $512\times 512$. $^*$computed using the \href{https://github.com/zhijian-liu/torchprofile}{torchprofile} package. }
    \label{tab:ade} \vspace{-4mm}
\end{table}

\begin{table*}[t]
    \centering
    %\vspace{-1mm}
    \resizebox{0.95\textwidth}{!}{
    \begin{tabular}{l|c|p{1.1cm}|p{1.1cm}|p{1.4cm}|p{1.1cm}|p{1.1cm}|l|l|l|l|l|l|p{1.1cm}|p{1.1cm}|l}
     \multirow{2}{*}{\#Clusters} & \multirow{2}{*}{Where?} & \multicolumn{3}{c|}{IN1K} & \multicolumn{8}{c|}{MS-COCO w/ Mask RCNN 1x} & \multicolumn{3}{c}{MIT-ADE20K w/ PaCa Head} \\ \cline{3-16}
            &  & \#Params  & FLOPs  & Top-1(\%) & \#Params  & FLOPs   & AP$^b$ & AP$^b_{50}$ & AP$^b_{75}$ & AP$^m$ & AP$^m_{50}$ & AP$^m_{75}$ & \#Params & FLOPs  & mIOU \\ \toprule          
          (100, 100, 100, 100) & \multirow{5}{*}{stage-wise} & 22.3 &  5.6 &  83.05 & 42.0 & 294 & \textbf{46.4} & 68.8 & 51.0 & \textbf{41.8} & 65.6 & 44.6 & 23.4 & 47.3& \textbf{48.3}\\ 
           (49, 64, 81, 100) & & 22.2 & 5.3  &  82.98 & 42.0 & 289 & 46.1 & 68.7 & 50.4 & 41.5 & 65.3 & 44.3 & 23.4 & 47.3 & 47.8  \\
           (100, 81, 64, 49) & & 22.2 & 5.3 &   82.87 & 42.0 & 291 & 46.1 & 68.4 & 50.3 & 41.7 & 65.4 & 44.7 & 23.7 & 47.3 & 47.6 \\ 
           (49, 49, 49, 0) & & 22.0  & 5.0  &  82.95 & 41.8 & 289 & 46.2 & 68.6 & 50.6 & 41.6 & 65.4& 44.3& 23.2 & 47.2 & 48.1\\ 
           (2, 2, 2, 0) & & 22.0  & 4.7  &  82.28 & 41.6 & 283 & 45.5 & 68.4 & 49.9 & 41.1 & 64.9 & 43.9 & 23.2 & 47.2 & 47.7 \\ 
          \textbf{(100, 100, 100, 0)} &  & 22.0 &  5.5 &  \textbf{83.08} & 41.8& 296 & \textbf{46.4} & 68.7 & 50.9 & \textbf{41.8} & 65.5 & 45.0 & 23.2 & 47.2 & \textbf{48.3}\\ \hline 
          (100, 100, 100, 100) & block-wise & 24.2 & 6.1 &  82.93 & 44.0 & 304 & 46.5 & 68.7 & 51.0 & 41.8 & 65.6 & 45.0 & 25.8 & 50.9 & 48.0\\ \bottomrule
        % MLP (Eqn.~\ref{eq:before_clustering_1}) & (100, 100, 100, 0) & \cmark & 22.6 & 5.9 & 83.13 & 42.4 & 303 & 46.6 & 69.0 & 51.3 & 41.9 & 65.7 & 45.0 & 24.0 & 50.0 & 48.2 \\ \bottomrule
    \end{tabular}
    }
    \caption{Ablation study on the number $M$ of clusters using the onsite clustering-via-convolution {\tt PaCa-Small} model on IN1K (Top-1), MS-COCO (with Mask-RCNN 1x) and MIT-ADE20k (with the proposed PaCa head). As mentioned in the submission (Sec.~\ref{sec:segmentation}), on MIT-ADE20k, the number of clusters in a stage of the ImageNet-pretrained backbone is reset to 200 if it is not zero. See details in text.}
    \label{tab:more_ablation_M} \vspace{-3mm}
\end{table*}

\subsection{Object Detection and Instance Segmentation} \label{sec:detection}\vspace{-2mm}

The challenging MS-COCO 2017 benchmark~\cite{lin2014microsoft} is used, which consists of a subset of train2017 (118k images) and a subset of val2017 (5k images). Following the common settings, we use the IN1K pretrained PaCa ViT models as the feature backbone, and test them using the Mask R-CNN framework~\cite{he2017mask} under the 1x schedule. 

\textbf{Accuracy.} Table~\ref{tab:coco} shows the comparisons. The proposed PaCa models obtain consistently better performance than other ViT variants. The clustering-via-MLP obtains slightly better performance than both the clustering-via-convolution and the external clustering with the small model configuration. With the base model configuration, the external clustering is slightly better than the clustering-via-convolution.    \textbf{Efficiency.} Overall, our PaCa models are significantly more efficient as shown by the FLOPs comparing with the baseline PVTv2. 

\subsection{Semantic Segmentation}\label{sec:segmentation} \vspace{-2mm}
The MIT-ADE20k~\cite{zhou2019semantic} benchmark is used, which is a challenging dense prediction task consisting of $L=150$ ground-truth classes. We use the IN1K pretrained PaCa ViT models as the feature backbone, and the proposed PaCa segmentation head network (Fig.~\ref{fig:paca_seg_head}). Since the pretrained PaCa backbones are trained with $M=100$ clusters that is smaller than $L$, we change $M=200$ in this task, and observe no issues of training, and better performance than the counterpart during our development.

Table~\ref{tab:ade} shows the comparisons. With the same UperNet~\cite{xiao2018unified} head,  the proposed PaCa-ViT backbones (stage-wise onsite clustering-via-convolution) consistently outperform other methods. With the proposed PaCa head, the proposed PaCa-ViT models further improve the performance, while significantly reducing the complexity compared with the UperNet head. This shows the effectiveness of the proposed PaCa head for semantic segmentation, which has a simple structure by design. It is also more effective than the semantic FPN~\cite{kirillov2019panoptic} head. With the PaCa segmentation head, the stage-wise onsite clustering-via-convolution models obtain better performance than the counterparts.

\subsection{Ablation Study}\label{sec:ablation}
\begin{figure}
    \centering
    \includegraphics[width=0.5\textwidth]{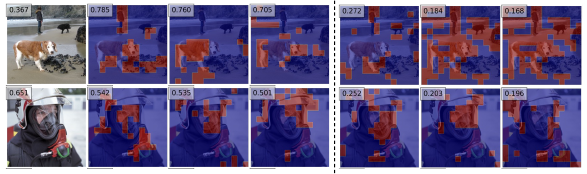}
    \caption{Examples of visualizing the clusters using the PaCa-Small model using the method presented in Sec.~\ref{sec:interpretability}. Both images are correctly classified by the model. The left three clusters are the top-3 in the positive group, and the right three clusters are the top-3 in the negative group. The top-k in either group is defined based on the prediction probability of the ground-truth class. For the first image, it is interesting to see the positive group leads to higher prediction probabilities than the raw input image. The full visualization is provided in the supplementary.}
    \label{fig:paca_xai} \vspace{-2mm}
\end{figure}

\vspace{-2mm}
In this section, we present an ablation study on the number $M$ of clusters in each of the four stages (Fig.~\ref{fig:PCAPVT}). 
The results are shown in Table~\ref{tab:more_ablation_M}. Interestingly, in terms of image classification performance, the number of clusters does not have a significant impact based on the cases tested (even with the number of clusters pushed to 2), which shows the robustness of the proposed PaCa models, but also suggests a potential improvement that may be worth exploring: Similar in spirit to the auxiliary loss used in the PaCa segmentation head (Fig.~\ref{fig:paca_seg_head}), some self-supervised loss functions (e.g., the loss function proposed in the Barlow Twins~\cite{zbontar2021barlow}) could be leveraged to induce learning diverse and complementary clusters for capturing underlying meaningful patterns (reusable and composable parts) at scene-/object-/part-levels. Based on the diversity of clusters, instance-sensitive cluster masks can be learned to filter out redundant clusters on the fly. Based on the visualization of learned clusters (Fig.~\ref{fig:paca_xai}), we observe redundant clusters and cluttered clusters. We leave those for the future work.

% \vspace{-2mm}
\section{Conclusion}\label{sec:conclusion}\vspace{-2mm}
This paper presents a patch-to-cluster attention (PaCa) module for learning efficient and interpretable Vision Transformers (ViTs). The proposed PaCa can address the quadratic complexity issue and account for the spatial redundancy of patches in the commonly used patch-to-patch attention. It also provides a forward explainer for diagnosing the explainability of ViTs. A simple learnable clustering module is introduced for easy integration in the ViT models. The proposed PaCa is also used in designing a lightweight yet effective semantic segmentation head network. In experiments, the proposed PaCa is tested in IN1K, MS-COCO and MIT-ADE20k benchmarks. It obtains consistently better performance than the prior art including the SWin-Transformers and PVTs. It also shows semantically meaningful qualitative results of the learned clusters. 
% The potential of the proposed PaCa beyond the tasks evaluated here and vision applications is discussed with promising preliminary experimental support. 

\section*{Acknowledgements} \vspace{-2mm}
{\small This research is partly supported by the Office of the Director of National Intelligence (ODNI), Intelligence Advanced Research Projects Activity (IARPA), via Contract No. 2021-21040700003,  ARO Grant W911NF1810295, NSF IIS-1909644, ARO Grant W911NF2210010, NSF IIS-1822477, NSF CMMI-2024688 and NSF IUSE-2013451. 
The views and conclusions contained herein are those of the authors and should not be interpreted as necessarily representing the official policies or endorsements, either expressed or implied, of the ODNI, IARPA, ARO, NSF, DHHS or the U.S. Government. The U.S. Government is authorized to reproduce and distribute reprints for Governmental purposes not withstanding any copyright annotation thereon. The authors are grateful for the constructive comments by anonymous reviewers and area chairs. }

%%%%%%%%% REFERENCES
{\small
\bibliographystyle{ieee_fullname}
\bibliography{egbib}
}

\clearpage
\newpage 

\appendix

\setcounter{figure}{6} %\renewcommand{\thefigure}{S.\arabic{figure}}
\setcounter{table}{4} %\renewcommand{\thetable}{S.\arabic{table}}

\section{Model Specifications}
\begin{figure*} [h]
    \centering
    \includegraphics[width=0.99\textwidth]{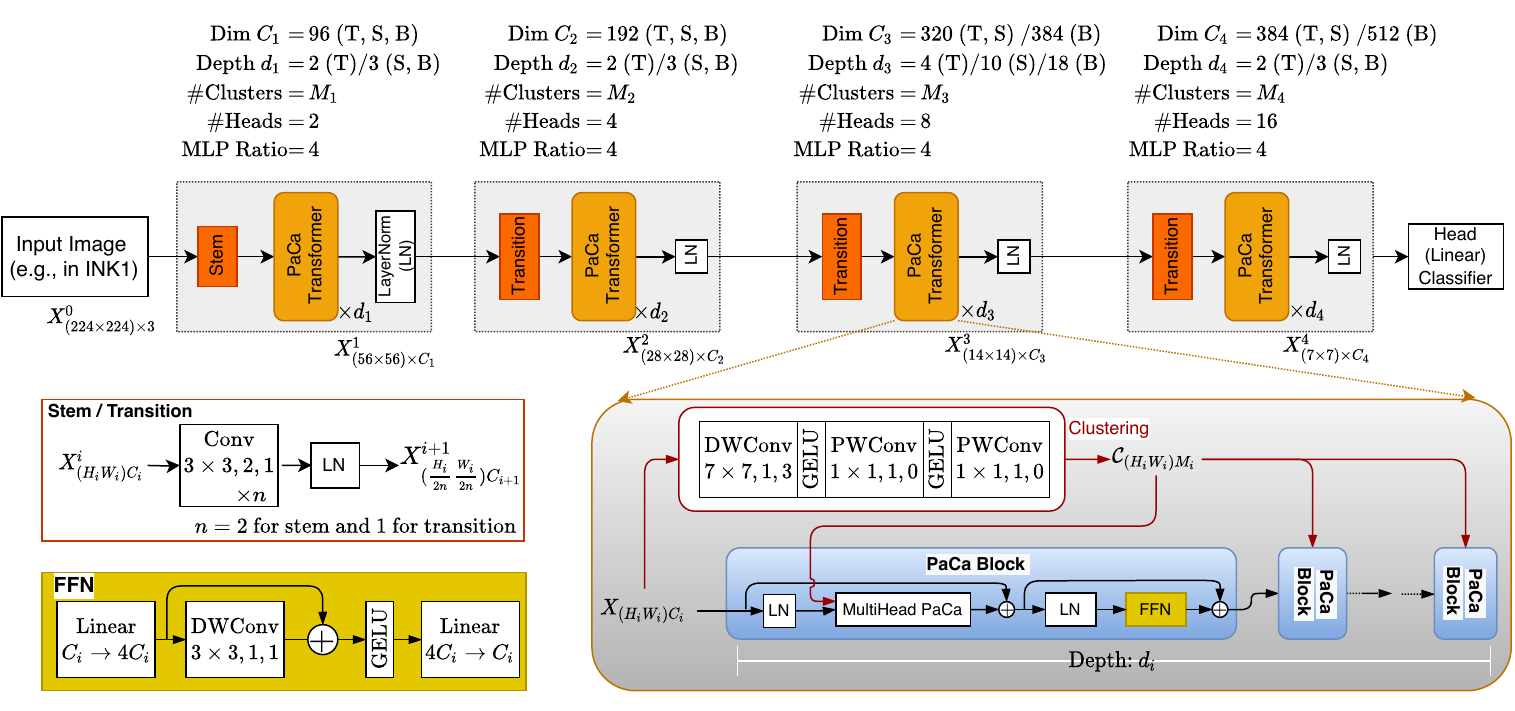}
        \caption{Mode specifications in the main experiments (elaborated on the Fig.~\ref{fig:PCAPVT} in the paper). We test three configurations: Tiny (T), Small (S) and Base (B). For the main results (see Tables~\ref{tab:in1k},~\ref{tab:coco} and ~\ref{tab:ade} in the paper), the number of clusters are $M_1=M_2=M_3=100$ and $M_4=0$ (i.e., degenerated back to the vanilla Transformer as done in the PVTv2~\cite{wang2021pvtv2}), and the cluster assignment $\mathcal{C}_{(H_iW_i)M_i}$ is \textit{shared} between all blocks in a stage as shown in the right-bottom. In the ablation studies, different configurations of the number of clusters at different stages are tested. A different clustering module based on a plain MLP is also tested (see Eqn.~\ref{eq:before_clustering_1} in the paper). The FFN implementation is adapted from the Inverted Residual Block proposed in the MobileNetv2~\cite{sandler2018mobilenetv2}, which is also used in PVTv2~\cite{wang2021pvtv2}. We add the shortcut connection over the depth-wise convolution to induce it to play the role of positional encoding more faithfully as proposed in~\cite{chu2021conditional}. }
    \label{fig:model_specification_main}
\end{figure*}

\begin{figure*} [h]
    \centering
    \includegraphics[width=0.99\textwidth]{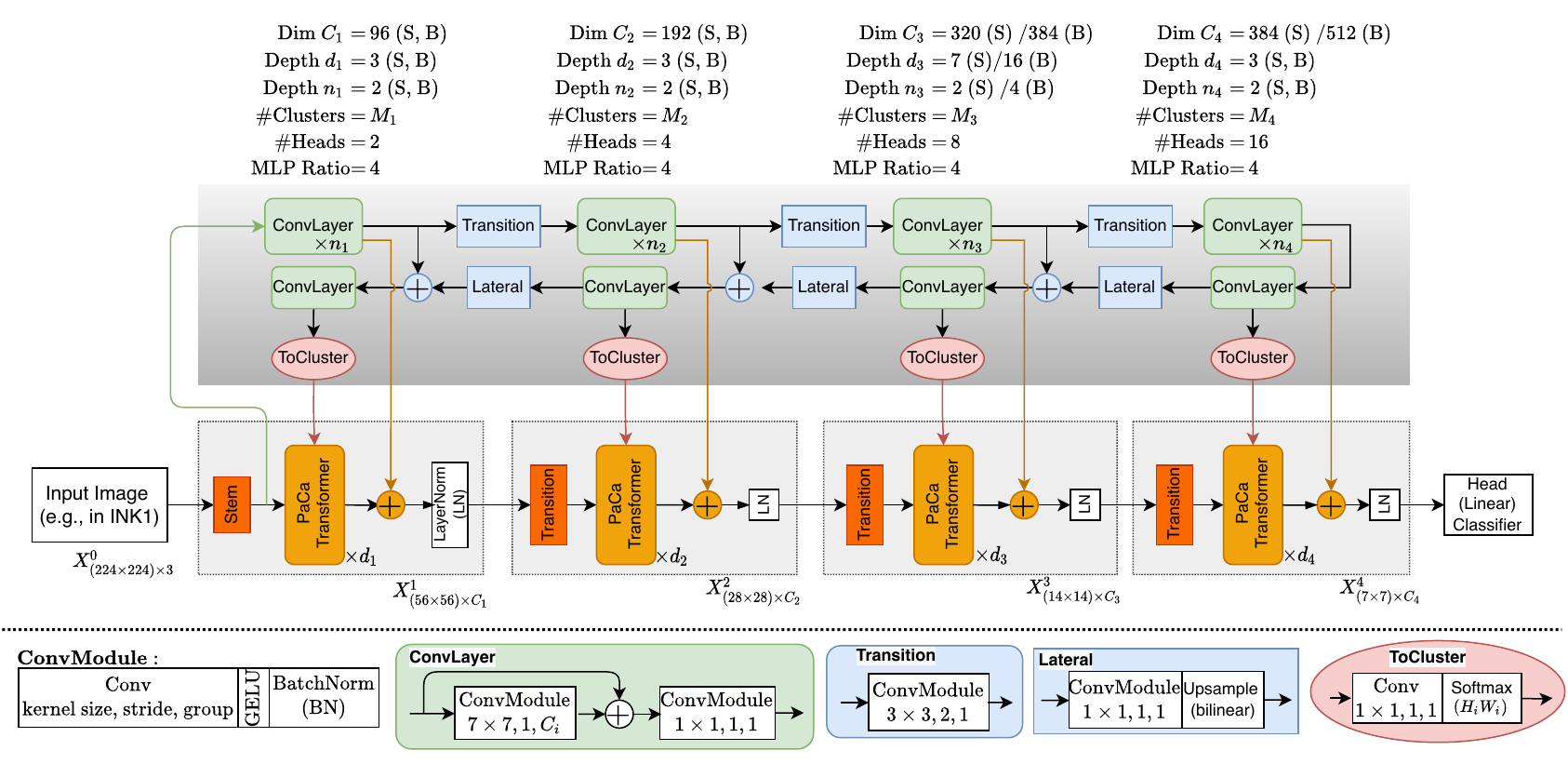}
    \caption{Mode specifications with external clustering teacher networks (elaborated on the Fig.~\ref{fig:paca_conv} in the paper). We test two configurations: Small (S) and Base (B). The ViT branch has the same specifications as shown in Fig.~\ref{fig:model_specification_main}.  In the experiments, the number of clusters are $M_1=M_2=M_3=M_4=100$, and the cluster assignment $\mathcal{C}_{(H_iW_i)M_i}$ from the teacher network is \textit{shared} between all blocks in a stage. The ``ConvLayer" module is adapted from the building block used in the ConvMixer~\cite{trockman2022patches}.  }
    \label{fig:model_specification_teacher}
\end{figure*}

We provide details of the model specifications shown in Fig.~\ref{fig:model_specification_main} (elaborated on the Fig.~\ref{fig:PCAPVT} in the paper) and Fig.~\ref{fig:model_specification_teacher} (elaborated on the Fig.~\ref{fig:paca_conv} in the paper). 

\section{Implementation Details}

\subsection{Experimental Details of Image Classification}
For image classification in the IN1K~\cite{deng2009imagenet}, all models in Sec.~\ref{sec:classification} are trained on the {\tt training set} for fair comparisons with the top-$1$ accuracy ($\%$) on the {\tt validation set}. The training receipt is adopted from DeiT~\cite{touvron2021training}, which has been widely used in training ViT variants. Table~\ref{tab:classification_recipe} shows the exact configurations used in our experiments. \textbf{Data Augmentation in Training:} we apply random cropping, random horizontal flipping~\cite{szegedy2015going}, label-smoothing regularization~\cite{szegedy2016rethinking}, mixup~\cite{zhang2017mixup}, and random erasing~\cite{zhong2020random} as data augmentations. During training, we employ AdamW~\cite{loshchilov2017decoupled} with a momentum of $0.9$, a mini-batch size of $128$, and a weight decay of $0.05$ to optimize models. The initial base learning rate is set to $5\times 10^{-4}$ and decreases following the cosine schedule~\cite{loshchilov2016sgdr}. The drop-path regularization is also used~\cite{huang2016deep}. All of our PaCa ViT models are trained for 300 epochs from scratch on 10 A100 GPUs with a learning rate auto-scaling heuristic method applied (see Table~\ref{tab:classification_recipe}). \textbf{Evaluation:} We apply a single center crop ($224\times 224$) on the validation set in evaluating the classification accuracy. We us the latest {\tt timm} package~\cite{rw2019timm}.

\begin{table}[t]
    \centering
    \resizebox{0.35\textwidth}{!}{
    \begin{tabular}{r|l} 
    Config. & Value \\ \toprule
         batch\_size & 128 \\
train\_interpolation & 'bicubic' \\
epochs &  300 \\ \midrule
opt &  'adamw' \\
opt\_eps &  1e-8 \\
opt\_betas & (0.9, 0.999) \\
momentum &  0.9 \\
weight\_decay &  0.05 \\
auto\_scale\_lr &  true \\
lr & 5e-4 \\
min\_lr & 5e-6 \\
sched &  'cosine' \\
warmup\_epochs &  5 \\
warmup\_lr &  5e-7  \\
cooldown\_epochs &  0 \\ 
amp &  True \\
clip\_grad & none (T, S) / 1.0 (B) \\ 
clip\_mode & norm \\ 
drop\_path\_rate & 0.1 (T, S) / 0.5 (B) \\ \midrule
color\_jitter & 0.4 \\
smoothing & 0.1 \\
reprob &  0.25 \\
remode & 'pixel' \\
recount &  1 \\
aa & 'rand-m9-mstd0.5-inc1' \\
mixup & 0.8 \\
cutmix &  1.0 \\
mixup\_prob &  1.0 \\
mixup\_switch\_prob &  0.5 \\
mixup\_mode &  'batch' \\ \bottomrule
    \end{tabular}}
    \caption{Training configurations used in training the proposed PaCa ViT models in IN1K following the {\tt timm} package~\cite{rw2019timm}. We train three model specifications: Tiny (T), Small (S) and Base (B). This training receipt is adapted from~\cite{touvron2021training} and  often applied and tuned for training with 8 GPUs. We use 10 GPUs to take the full advantage of the server we have and to speed up the experiments. Accordingly, we apply a heuristic ``auto\_scale\_lr" setting which scales ``lr", ``min\_lr" and ``warmup\_lr" in this table with the factor ``batch\_size $\times$ nb\_gpus / 512" (i.e., $2.25$ in our settings) to account for the increased number of total images per batch with 10 GPUs used. We note that scaling these learning rate related hyperparamters often has slightly negative effects on performance.}
    \label{tab:classification_recipe}
\end{table}

\subsection{Experimental Details of Object Detection and Instance Segmentation}

We use the proposed PaCa ViT models (Tiny, Small and Base) as the feature backbones in the Mask R-CNN~\cite{he2017mask} and test them on the MS-COCO~\cite{lin2014microsoft} dataset. All models in Sec.~\ref{sec:detection} are trained on MS-COCO {\tt train2017} (118k images) and evaluated on {\tt val2017} (5k images). We use the MMDetection~\cite{chen2019mmdetection} package (version 2.25.2) in experiments.  We apply the weights pre-trained on IN1K to initialize the backbone and Xavier~\cite{glorot2010understanding} in initializing the remaining layers in the Mask R-CNN (the default in the MMDetection). We adopt the 1x schedule in training (i.e., 12 epochs used in training). %Following the Swin-Transformer~\cite{liu2021swin}, we utilize the same multi-scale training strategy by randomly resizing the shorter side of an image within the range of $[480, 800]$ , while the longer side does not exceed $1,333$ pixels. 
In both training and evaluation, the shorter side of the input image is fixed to $800$ pixels with the longer side retained not exceeding $1,333$ pixels. We train Mask R-CNN with our PaCa ViT backbones using batch size 16 on 8 A100 GPUs (i.e., 2 images per GPU)~\footnote{We follow the provided recipes and do not apply the auto-scaling heuristic to take advantage of the 10 GPUs we have on the server (that is done for IN1K training, see Table~\ref{tab:classification_recipe}), since we observe the auto-scaling heuristic has more significantly negative impacts on performance on the downstream tasks and the training on the downstream tasks consumes much less time than that in IN1K.}, following the recipes in the MMDetection package which use the AdamW~\cite{loshchilov2017decoupled} optimizer with an initial learning rate of $1\times 10^{-4}$, and a weight decay 0.05. The parameters of the normalization layers are excluded from the weight decay.  

\subsection{Experimental Details of Image Semantic Segmentation}
We use the proposed PaCa ViT models (Tiny, Small and Base) as the feature backbones and two different segmentation head sub-networks, the UpperNet~\cite{xiao2018unified} and our proposed PaCa segmentation head (Sec.~\ref{sec:paca_seg_head}).  We test them on the MIT-ADE20k~\cite{zhou2019semantic} dataset. In training, we randomly resize and crop images to the resolution of $512\times 512$. In evaluation, images are resized to have a shorter side of $512$ pixels. The longer side is fixed not to exceed $2,048$ pixels. We use the MMSegmentation~\cite{mmseg2020} package (version 0.29.0) in experiments. 
 We apply the weights pre-trained on IN1K to initialize the backbone and Xavier~\cite{glorot2010understanding} in initializing the head sub-network (the default in the MMSegmentation). We train our PaCa models with 160k iterations using batch size 16 on 8 A100 GPUs (i.e., 2 images per GPU). We adopt the default recipes provided in the MMSegmentation package, using the AdamW~\cite{loshchilov2017decoupled} optimizer with an initial learning rate of $6\times 10^{-5}$ for the backbone, and $6\times 10^{-4}$ for the head sub-network, and a weight decay 0.01. The parameters of the normalization layers are excluded from the weight decay. As mentioned in Sec.~\ref{sec:segmentation}, we increase the number of clusters used in the backbone from $100$ to $200$ to account for the increased number of ground-truth classes in the MIT-ADE20k (150 classes). Due to this change, we set the initial learning rate to $6\times 10^{-4}$, the same as the head sub-network, for the clustering layer (Eqn.~\ref{eq:clustering}).

% \section{Some More Results and Analyses}
% Table~\ref{tab:more_ablation_M} and Table~\ref{tab:more_ablation_teacher} show more detailed comparisons in the ablation studies of the number of clusters and the external clustering assignment teacher network. For the number of clusters, the simple setting ($M_1=M_2=M_3=100$ and $M_4=0$) used in the main experiments in the submission is an overall better one. For the external clustering assignment teacher networks, their performance on downstream tasks are not consistently better. It may related to the the design of the teacher network. The current design (Fig.~\ref{fig:model_specification_teacher}) has BatchNorm in the ConvModule, adopted from the ConvMixer~\cite{trockman2022patches}. They are frozen in finetuning on the downstream tasks. In the future, we plan to investigate to leverage the ConvNeXt~\cite{liu2022convnet} design. 

\section{Examples of Learned Clusters}

\begin{figure*}
    \centering
    \includegraphics[width=0.99\textwidth]{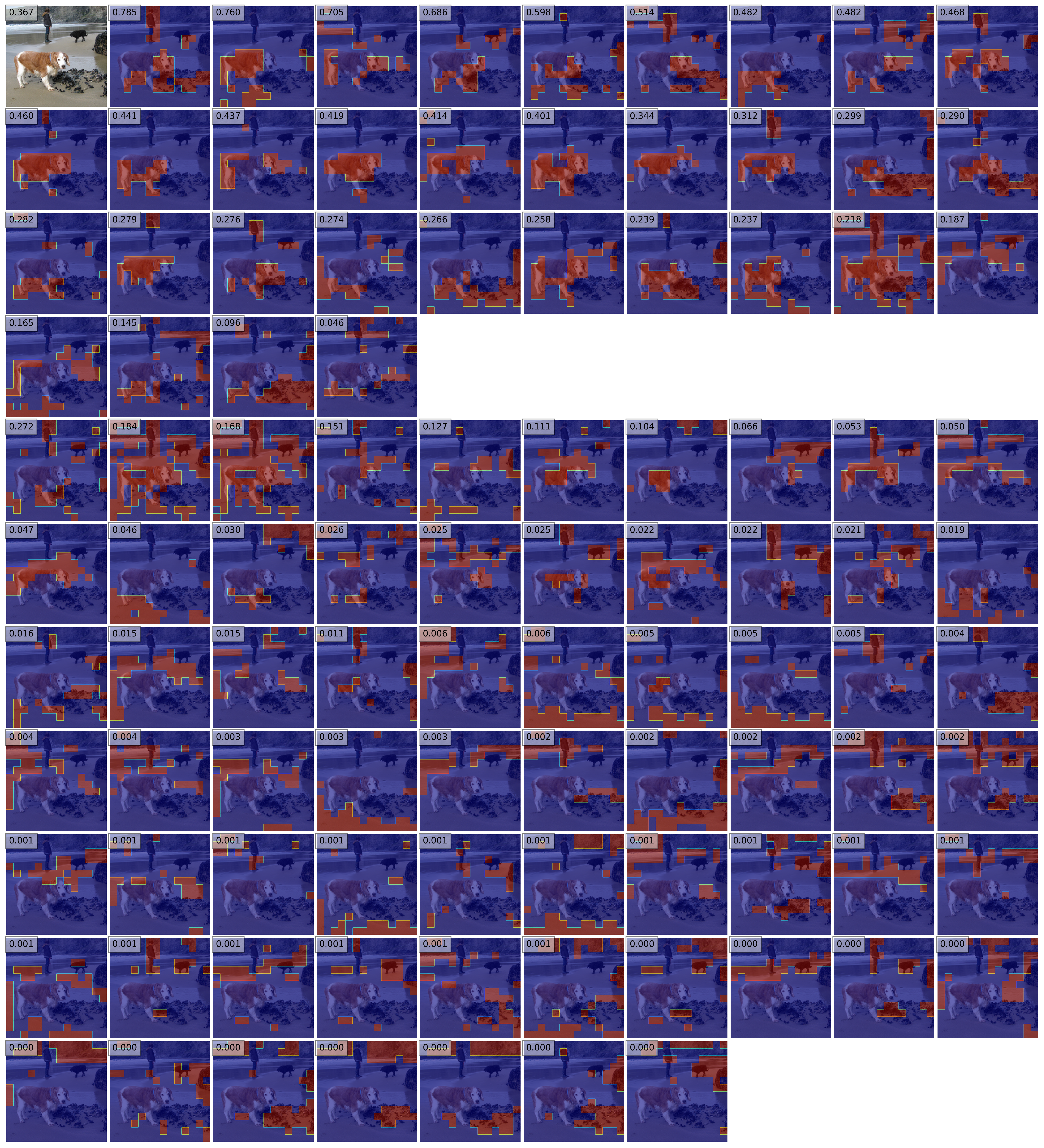}
    \caption{Visualizing the learned clusters with an image (id: 10933) in the IN1K validation set. We use the PaCa-Small network (Table~\ref{tab:in1k}). This image is correctly classified by the model. The 100 clusters learned at the third stage are used. The left-top image is the input image with the original predicted probability for the ground-truth class shown in the left-top box. The first 4 rows show the masked images in the positive group. It is interesting to see that many masked images can lead to higher predicted probabilities for the ground-truth class. The remaining rows show the masked images in the negative group. Although the first several images in the negative group have the predicted probabilities larger than some images in the positive group, the ground-truth class is not the top-1. }
    \label{fig:10933}
\end{figure*}

\begin{figure*}
    \centering
    \includegraphics[width=0.99\textwidth]{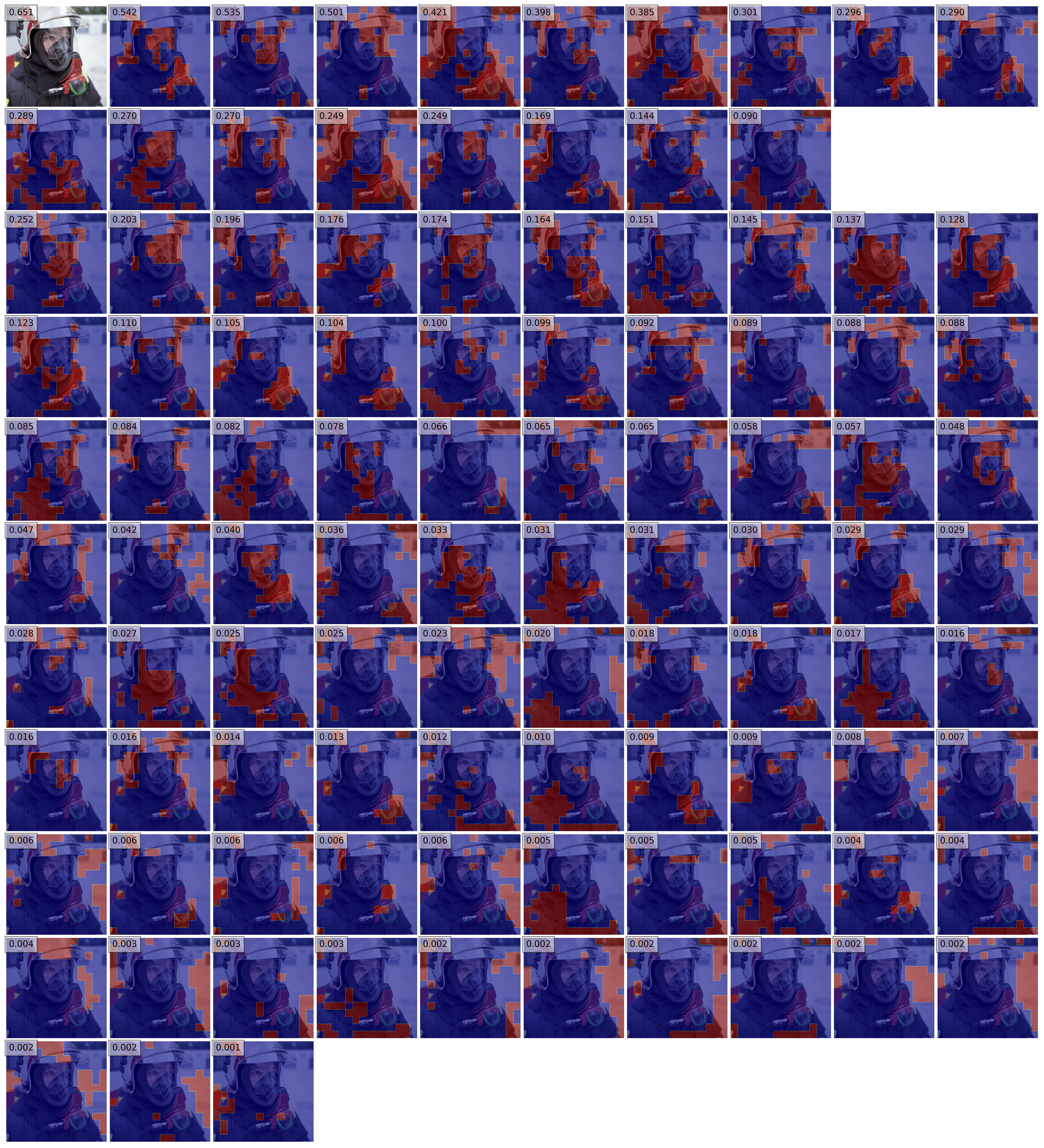}
    \caption{Visualizing the learned clusters with an image (id: 34561) in the IN1K validation set. We use the PaCa-Small network (Table~\ref{tab:in1k}). This image is correctly classified by the model. The 100 clusters learned at the third stage are used. The left-top image is the input image with the original predicted probability for the ground-truth class shown in the left-top box. The first 2 rows show the masked images in the positive group. For this examples, all the masked images have smaller predicted probabilities than the original unmaimage.  The remaining rows show the masked images in the negative group. Although the first several images in the negative group have the predicted probabilities larger than some images in the positive group, the ground-truth class is not the top-1. }
    \label{fig:34561}
\end{figure*}

We show all the clusters elaborating Fig.~\ref{fig:paca_xai} in the paper in Figures~\ref{fig:10933} and ~\ref{fig:34561}.

\end{document}